\let\oldvec\vec
\documentclass[runningheads]{llncs}
\let\vec\oldvec
\usepackage[english]{babel}
\usepackage{amsfonts}
\usepackage{amsmath}
\usepackage{blindtext}
\usepackage{graphicx}
\usepackage{marvosym}
\usepackage[lofdepth,lotdepth]{subfig}
\usepackage{soul}
\usepackage[dvipsnames]{xcolor}
\usepackage{bbm}

\newenvironment{system}[1][1.1]
 {\renewcommand{\arraystretch}{#1}
  \left\{\begin{array}{@{}l@{}}}
 {\end{array}\right.\kern-\nulldelimiterspace}

\newcommand{\R}{\mathbb{R}}
\newcommand{\eps}{\varepsilon}
\newcommand{\Ha}{\mathcal{H}}
\newcommand{\E}{\mathbb{E}}
\newcommand{\one}{\mathbbm{1}}

\begin{document}
\title{Measuring shape relations using $r$-parallel sets\thanks{Supported by the Center for Stochastic Geometry and Advanced Bioimaging.}}
%
%
\author{Hans JT Stephensen\inst{1}\textsuperscript{(\Letter)} \and Anne Marie Svane\inst{2} \and
Carlos Benitez\inst{1,3} \and
Steven A. Goldman\inst{1,3} \and
Jon Sporring\inst{1}}

\authorrunning{Stephensen et al.}
%
\institute{
University of Copenhagen, Copenhagen, Denmark\\ \email{hast@di.ku.dk} \and  Aalborg University, Aalborg, Denmark \and
University of Rochester, Rochester, USA
}
%
\maketitle              
\begin{abstract}
Geometrical measurements of biological objects form the basis of many quantitative analyses.  Hausdorff measures such as the volume and the area of objects are simple and popular descriptors of individual objects, however, for most biological processes, the interaction between objects cannot be ignored, and the shape and function of neighboring objects are mutually influential.

In this paper, we present a theory on the geometrical interaction between objects based on the theory of spatial point processes. Our theory is based on the relation between two objects: a reference and an observed object. We generate the $r$-parallel sets of the reference object, we calculate the intersection between the $r$-parallel sets and the observed object, and we define measures on these intersections. Our measures are simple like the volume and area of an object, but describe further details about the shape of individual objects and their pairwise geometrical relation. Finally, we propose a summary statistics for collections of shapes and their interaction.

We evaluate these measures on a publicly available FIB-SEM 3D data set of an adult rodent.




\end{abstract}
\section{Introduction}

Measuring the geometry and statistics of objects is a fundamental tool used in all areas of the natural sciences. Geometric object-descriptors vary in complexity from simple measures of point counts, area, and volume to parameterized domain-specific shape models, see \cite{Zhang2004} for a review of shape representations.

In many cases, we are further interested in the relation between objects to answer questions like: How do synaptic vesicles distribute in the neighborhood of a synapse during stress~\cite{Khanmohammadi2015}? How are astrocytes distributed w.r.t. the position and shapes of their nearby neuronal cells in amyotrophic lateral sclerosis~\cite{medvedev2014glia}? What is the relation of the position and shape of the cartilage of the tibia and femur and osteoarthritis~\cite{Marques2013}? A simple approach is to summarize each object as a point and consider the set of points as a point process that has a well-developed theory and readily available software, e.g.,~\cite{baddeley15}.


In this article, we adopt the perspective of the cross $K$-functions from the statistics of point processes, which describe the relation between paired point processes \cite{dixon2014r}.
A classic example use of the cross $K$-function is to model the occurrence of crime and the locations of police stations by point processes $\mathcal{X}$ and $\mathcal{Y}$, respectively.
In the cross $K$-function setting, we measure the expected number of crime occurrences $x\in \mathcal{X}$ within distance $r$ of a police station $y\in \mathcal{Y}$, i.e., for discrete point sets $\mathcal{X},\mathcal{Y}\subset\R^2$
\begin{equation}
    K(r) = \E_{y\in\mathcal{Y}}\big|\{x|d(x,y)\leq r, x\in\mathcal{X}\}\big| \enspace ,
    \label{eq:crossK}
\end{equation}
where $d$ is a distance measure, often the Euclidean distance, and $|\cdot|$ is the set-size operator, and $\E$ is the conditional expectation given $y\in\mathcal{Y}$.

\begin{figure}
\centering
\includegraphics[trim={40pt 0 40pt 0},clip, width=0.35\textwidth]{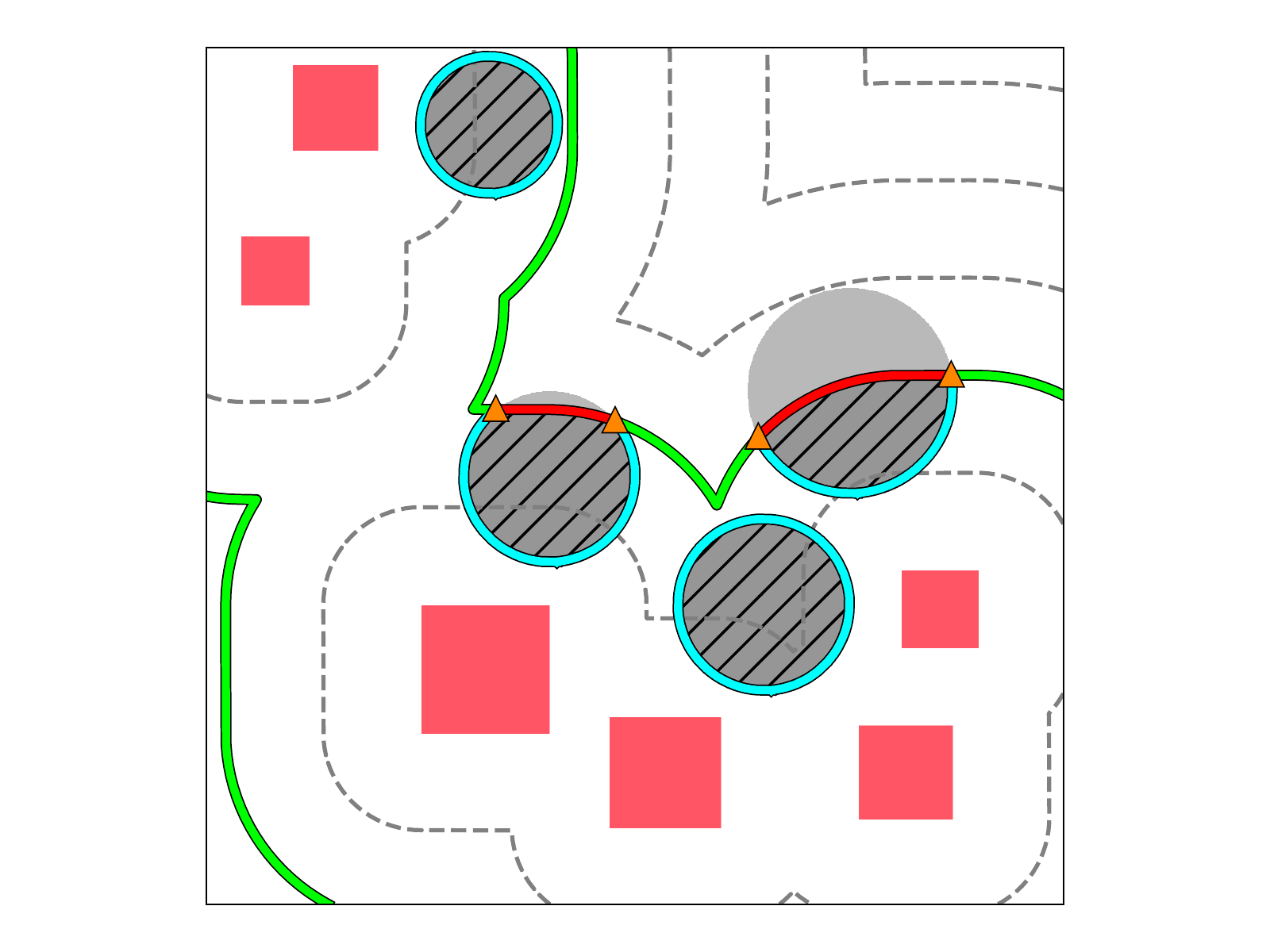}
\caption{An example of how the measures look for some simple 2D shapes. The disks are the observed objects and the squares are the reference objects. The green line shows the boundary of the r-parallel, giving rise to one counting measure (orange triangles), two length measures (red and blue contours) and one area measure (hatched grey region).}
\label{fig:fig_ex}
\end{figure}

Here, we extend the cross $K$-function to general geometric objects. We will consider objects $X, Y \subseteq \R^d$, which we will call the observed and the reference objects, and these objects may be points but also surfaces and solids. We extend the notion of distance as the shortest distance between two objects. As an example, in Fig.~\ref{fig:fig_ex} we show equidistant curves from the red reference objects and how these distance curves interact with the blue observed objects.
To quantify the observed objects w.r.t.\ the reference objects, we calculate a family of sets $Y^r$ as all points within a distance $r$ from $Y$, which we call the $r$-parallel sets. For each $r$-parallel, we can measure the number of points, the length of the inner and outer contours, and area of the intersection of objects with the $r$-parallel. These are also known as Hausdorff measures of the intersection between $Y^r$ and $X$.

A few published studies use aspects of this method \cite{medvedev2014glia,gavrilov2018astrocytic}, but the formality and theoretical foundation have been absent so far. Our contribution is to remedy this shortcoming and broaden the scope to include the full family of $n$-dimensional Hausdorff measures. To demonstrate the usefulness of our method, we present a segmentation of a publicly available FIB-SEM 3D dataset of an adult rodent~\cite{lausanne} in greater detail than previously available and use it as a subject for analysis using our proposed measures.
%
%
%
\section{Method Description}

We will measure the shape and position of some observed object $X\subseteq \R^d$ with respect to a reference object $Y\subseteq \R^d$.
For $r>0$, we measure the part of $X$ that lies within distance $r$ from $Y$.
The set of all points within distance $r$ from $Y$ is called the  \emph{$r$-parallel set} of $Y$ and is given by
\begin{equation}
Y^r = \{\alpha\in \R^d \mid \inf_{y\in Y}d(\alpha,y) \leq r\} \enspace .
\end{equation}
Here $d(\alpha, y)$ denotes the distance between $\alpha$ and $y$, typically the Euclidean distance. The points in $X$ having at most distance $r$ from $Y$ is thus the intersection $X\cap Y^r$.

We measure $X \cap Y^r$ by a measure $\mu(X, Y^r)$.
In all our applications, $\mu$ has the form
\begin{equation}
\mu_{\eps,\eps'}(X,Y^r) = \Ha^{d-\eps -\eps'} (\partial^\eps X \cap \partial^{\eps'}Y^r) \enspace ,
\end{equation}
where $\Ha^k$ denotes the $n$-dimensional Hausdorff measure, $\eps,\eps'\in \{0,1\}$, and for a closed set $C\subseteq \R^d$, $\partial^0 C=C$ is just $C$ itself, and $\partial^1 C =\partial C$ is the boundary of $C$. The interpretation of $\Ha^k$ and $\mu_{\eps,\eps'}(X,Y^r)$ in 2D and 3D is shown in Table~\ref{table2D3D}.

\begin{table}
\begin{center}
\begin{tabular}{ccccc}
	\hline
	$d=2$ & $(\eps,\eps')$ & $\Ha^{d-\eps-\eps'}$ & $\partial^\eps X \cap \partial^{\eps'} Y^r$ & Interpretation  of $\mu_{\eps,\eps'}(X,Y^r)$ \\ \hline
	 & $(0,0)$ & Area & $X  \cap  Y^r $&  Area of cut\\
	 & $(0,1)$ & Curve length & $ X \cap \partial Y^r$ & Boundary length of cut in interior of $X$\\
	 & $(1,0)$ & Curve length & $\partial X \cap Y^r$ & Boundary length of cut in boundary of $X$\\
	 & $(1,1)$ & Point counts & $ \partial X \cap \partial Y^r$ &  Point count in boundary intersection\\
	\hline
    \hline
    $d=3$ & $(\eps,\eps')$ & $\Ha^{d-\eps-\eps'}$ & $\partial^\eps X \cap \partial^{\eps'} Y^r$ & Interpretation of $\mu_{\eps,\eps'}(X,Y^r)$ \\ \hline
     & $(0,0)$ & Volume & $X \cap Y^r$&  Volume of cut\\
     & $(0,1)$ & Surface area & $ X \cap \partial Y^r $&Surface area of cut in interior of $X$\\
     & $(1,0)$ & Surface area & $\partial X \cap Y^r$ & Surface area of cut in boundary of $X$\\
     & $(1,1)$ & Curve length & $ \partial X \cap \partial Y^r$ &  Length of boundary intersection\\
    \hline
\end{tabular}
\end{center}
\caption{Interpretation  of $\mu_{\eps,\eps'}$ in 2D and 3D. The term cut refers to $ X \cap Y^r$.}
\label{table2D3D}
\end{table}

Since the boundary of a boundary is the empty set, i.e., $\partial\partial\cdot = \emptyset$, this is the complete set of measures in 2 and 3 dimensions, and these lists generalize naturally to any dimension. Further, $\mu_{\eps,\eps'}(X,Y^r)$ is continuous in $r$, and under mild conditions on the shapes $X$ and $Y$,
\begin{equation}
    \frac{d}{dr}\mu_{0,0}(X,Y^r) = \mu_{0,1}(X,Y^r) \enspace ,
    \label{eq:derivative}
\end{equation}
for almost all values of $r$. That is, the area of the part of the surface of $\partial Y^r$ lying inside $X$ gives the instantaneous change in the volume of the intersection of $X$ and $Y^r$. However, in general $ \frac{d}{dr}\mu_{1,0}(X,Y^r) \neq \mu_{1,1}(X,Y^r)$.


To model a collection of objects, we equip each object with a reference point and obtain a marked point process $\mathcal{X}=\{x_i,X_i\}_{i\geq 0} $ on $\R^d \times \mathcal{C}$, where the mark space $\mathcal{C}$ is the space of all compact sets in $\R^d$ with smooth boundary. The point $x_i$ can be thought of as the location of the object and the associated mark $X_i$ as the shape. This is also known as a \emph{germ-grain process} \cite{weil}. The collection of objects, sometimes called the \emph{germ-grain model}, is then
\begin{equation}
    \bigcup_{i\geq 0} (x_i + X_i) \enspace ,
\end{equation}
where $v+X = \{v+x\mid x\in X\}$.

Let $\mathcal{X}, \mathcal{Y} $ be germ-grain processes modelling the observed and reference objects, respectively.  We assume that the processes are jointly stationary. Writing $v + \mathcal{X}  = v + \{x_i,X_i\}_{i\geq 0}  = \{v + x_i,X_i\}_{i\geq 0}$, stationarity means that $(v+\mathcal{X} ,v+ \mathcal{Y} )$ has the same distribution as $(\mathcal{X} , \mathcal{Y})$ for any translation vector $v\in \R^d$.

A global functional summary statistic is given by sampling reference particles in an observation window $W$ and summing $\mu_{\eps,\eps'}$ for each pair of a sampled reference object and an observed object:
\begin{equation}
\label{eq:summary_1}
\hat{K}_{\eps,\eps'}(r)= \frac{1}{\Ha^d(W)\rho_{\mathcal{X}}\rho_{\mathcal{Y}}} \sum_{(y,Y)\in \mathcal{Y}} \one_{\{y\in W\}} \sum_{(x,X)\in \mathcal{X}}  \mu_{\eps,\eps'}( x+X , y+ Y^r) \enspace .
\end{equation}
Here $\rho_{\mathcal{X}}$ and $\rho_{\mathcal{Y}}$ are the spatial intensities of the point processes underlying $\mathcal{X}$ and $\mathcal{Y}$, respectively, and $\Ha^d(W)$ is the volume of the sampling window. Note that in order to compute $\hat{K}_{\eps,\eps'}(r)$, we must be able to observe $\mathcal{X}$ in a slightly larger window.

The expected value of $\hat{K}_{\eps,\eps'}$ is
\begin{equation}
\label{eq:summary_2}
{K}_{\eps,\eps'}(r)= \frac{1}{\Ha^d(W)\rho_{\mathcal{X}} \rho_{\mathcal{Y}}} \E\sum_{(y,Y)\in \mathcal{Y}} \one_{\{ y\in W\}} \sum_{(x,X)\in \mathcal{X}} \mu_{\eps,\eps'}( x+X ,y+ Y^r) \enspace .
\end{equation}
Due to the stationarity assumption, this is independent of the choice of sampling window.

As \cite{Ripley1979}, we also consider normalized functions,
\begin{align}
    N_{\eps'}(Y^r) &= \Ha^{d-\eps'} (\partial^{\eps'} Y^r \cap W) \enspace ,\label{eq:normalization_term}\\
    \nu_{\eps,\eps'}(X,Y^r) &= \frac{\mu_{\eps,\eps'}(X,Y^r)}{N_{\eps'}(Y^r)} \enspace ,
\end{align}
where $N_{\eps'}(Y^r)$ is the size of the $r$-parallel set. To get normalized summary statistics, we replace $\mu_{\eps,\eps'}$ by $\nu_{\eps,\eps'}$ in (\ref{eq:summary_1}) and (\ref{eq:summary_2}) getting $\hat{L}_{\eps,\eps'}$ and $L_{\eps,\eps'}$ resp.


\subsection{Examples of simple object relations}
\label{sec:analytical}
For simple objects, $\mu_{\eps,\eps'}$ may be evaluated analytically. Consider an infinite line/plane as reference object and a circle/sphere of radius $R$ in $\Omega$ and with $W=\Omega$ as observed objects, and further, consider the center of the circle/sphere to be a distance of $R$ from the line/plane, then for $r\geq 0$ and using the equations for a circular segment and spherical cap respectively,
\\[\abovedisplayskip]
{\setlength{\abovedisplayskip}{0pt}
\begin{tabular}{c|l|l}
  &  $\,\Omega=\R^2$, $\theta=2\arccos(1-r/R)$ & $\,\Omega=\R^3$\\\hline
  $\mu_{00}=$
  & {\begin{minipage}{0.42\textwidth} 
      \begin{flalign}
        &\begin{system}
          \frac{R^2}{2}(\theta-\sin\theta)\text{, if $r<2R$}\\
          \pi R^2\text{, otherwise}
        \end{system}&
      \end{flalign}
    \end{minipage}}
                                               & {\begin{minipage}{0.48\textwidth}
                                                   \begin{flalign}
                                                     &\begin{system}
                                                       \frac{\pi r^2(3R-r)}{3}\text{, if $r<2R$}\\
                                                       \frac{4\pi R^3}{3}\text{, otherwise}
                                                     \end{system}&
                                                   \end{flalign}
                                                 \end{minipage}}\\
  $\mu_{01}=$
  & {\begin{minipage}{0.42\textwidth}
      \begin{flalign}
        &\begin{system}
          2 R\sin\frac{\theta}{2}\text{, if $r<2R$}\\
          0\text{, otherwise}
        \end{system}&
      \end{flalign}
    \end{minipage}}
                                               & {\begin{minipage}{0.48\textwidth}
                                                   \begin{flalign}
                                                     &\begin{system}
                                                       \pi r(2R-r)\text{, if $r<2R$}\\
                                                       0\text{, otherwise}
                                                     \end{system}&
                                                   \end{flalign}
                                                 \end{minipage}}\\
  $\mu_{10}=$
  & {\begin{minipage}{0.42\textwidth}
      \begin{flalign}
        &\begin{system}
          \theta R\text{, if $r<2R$}\\
          2\pi R\text{, otherwise}
        \end{system}&
      \end{flalign}
    \end{minipage}}
                                               & {\begin{minipage}{0.48\textwidth}
                                                   \begin{flalign}
                                                     &\begin{system}
                                                       2\pi R r\text{, if $r<2R$}\\
                                                       4\pi R^2\text{, otherwise}
                                                     \end{system}&
                                                   \end{flalign}
                                                 \end{minipage}}\\
  $\mu_{11}=$
  & {\begin{minipage}{0.42\textwidth}
      \begin{flalign}
        &\begin{system}
          1\text{, if $r=0$ or $r=2R$}\\
          2\text{, if $0<r<2R$}\\
          0\text{, otherwise}
        \end{system}&
      \end{flalign}
    \end{minipage}}
                                               & {\begin{minipage}{0.48\textwidth}
                                                   \begin{flalign}
                                                     &\begin{system}
                                                       0\text{, if $r=0$ or $r=2R$}\\
                                                       2\pi\sqrt{2Rr - r^2}\text{, if $0<r<2R$}\\
                                                       0\text{, otherwise}
                                                     \end{system}&
                                                   \end{flalign}
                                                 \end{minipage}}
\end{tabular}}\\[\belowdisplayskip]
%
For this example, $N_{\eps'}=\infty$, hence the normalized functions are all 0.

\subsection{Implementation details}
Our algorithm takes a triangulated surface-mesh of objects. Firstly, the shortest distance to $Y$ is sampled on a regular grid in $W$. This we call the distance map $D$. For $\mu_{0,\eps'}$ and related functions, we extend the surface-mesh of the observed object $X$ with a tessellation of their interior into a collection of $d$-dimensional simplexes.
For each vertex in $X$ or its extension, the shortest distance to $Y$ is estimated as a linear interpolation into $D$. For each $r$-parallel set $Y^r$, we identify interior, intersecting, and exterior simplexes in $X$. For the intersecting simplexes, we estimate the intersecting line/surface by linear interpolation. For $\mu_{1,0}$ and $\mu_{0,1}$, we sum the surface of intersections, for $\mu_{0,0}$, we sum the area/volume of the interior simplexes the relevant part of the intersecting simplexes. For $\mu_{1,1}$ and $d=2$, we count the number of intersections, and for $d=3$ calculate the length of the intersecting line.

\section{Experiments on synthetic objects}
In the following, we will give examples of experiments conducted on synthetic data.

\begin{figure}
\centering
\subfloat[$\mu_{00}$]{
\includegraphics[trim={0pt 0pt 0pt 0pt},clip, width=0.23\textwidth]{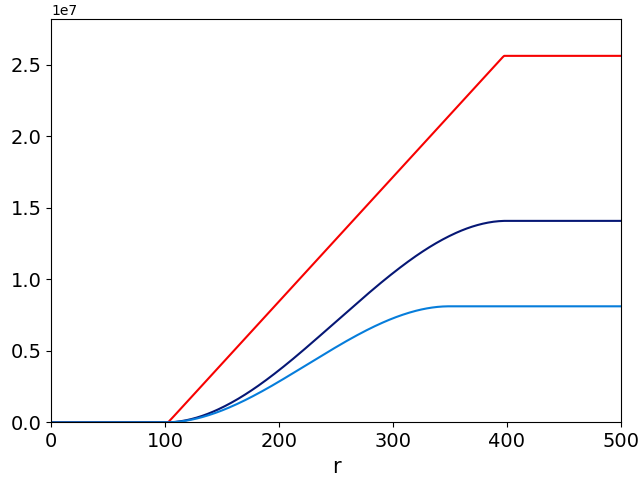}
\label{fig:subfig_sim_ex_4}}
\subfloat[$\mu_{01}$]{
\includegraphics[trim={0pt 0pt 0pt 0pt},clip, width=0.23\textwidth]{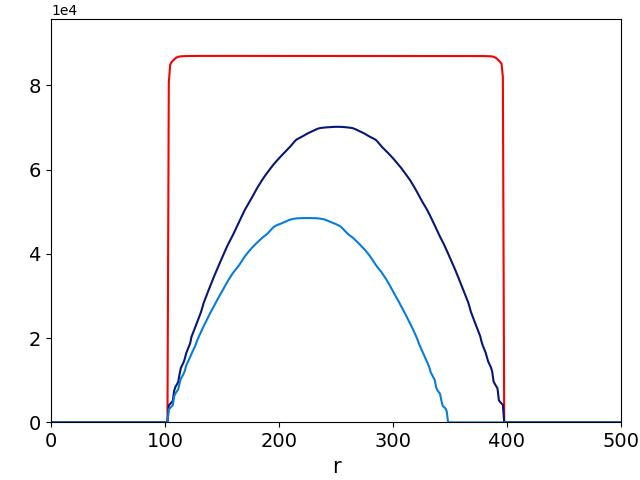}
\label{fig:subfig_sim_ex_2}}
\subfloat[$\mu_{10}$]{
\includegraphics[trim={0pt 0pt 0pt 0pt},clip, width=0.23\textwidth]{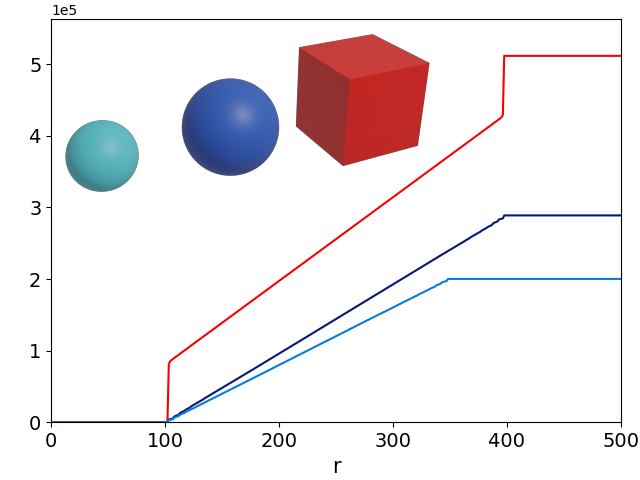}
\label{fig:subfig_sim_ex_3}}
\subfloat[$\mu_{11}$]{
\includegraphics[trim={0pt 0pt 0pt 0pt},clip, width=0.23\textwidth]{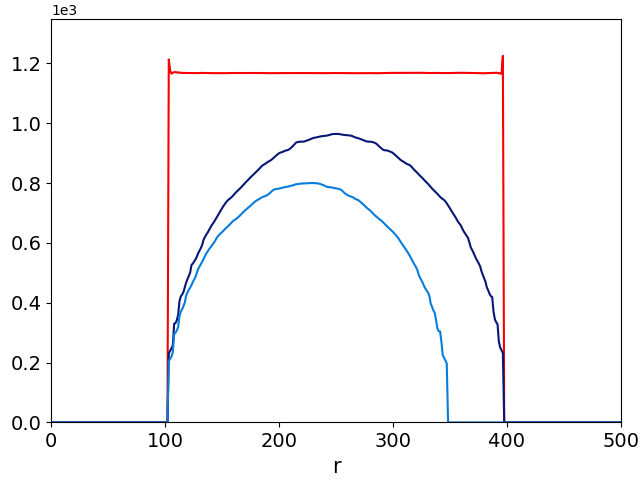}
\label{fig:subfig_sim_ex_1}}
\\
\subfloat[$\nu_{00}$]{
\includegraphics[trim={0pt 0pt 0pt 0pt},clip, width=0.23\textwidth]{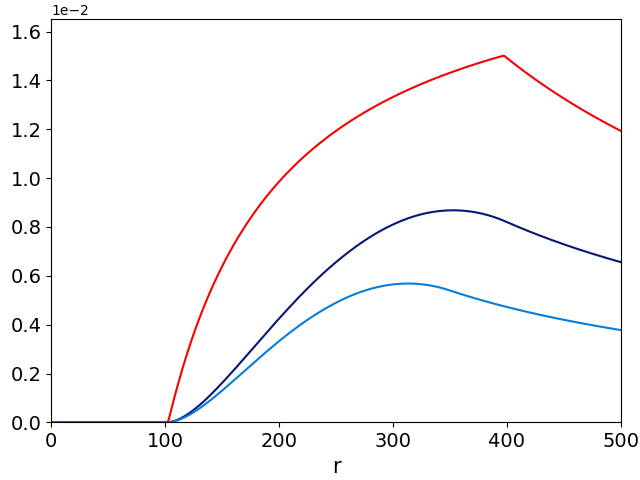}
\label{fig:subfig_sim_normed_ex_4}}
\subfloat[$\nu_{01}$.]{
\includegraphics[trim={0pt 0pt 0pt 0pt},clip, width=0.23\textwidth]{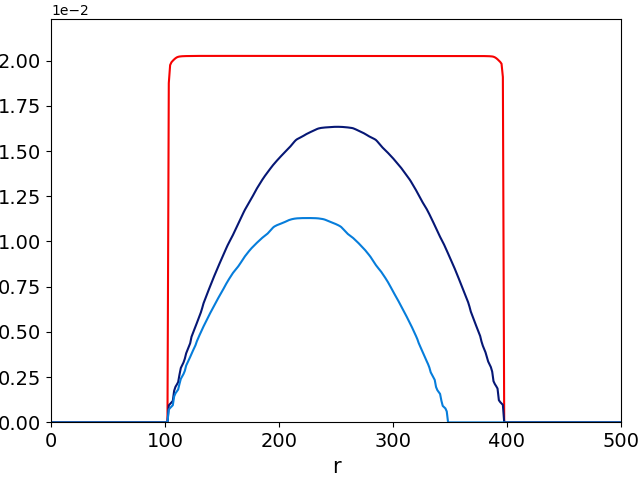}
\label{fig:subfig_sim_normed_ex_2}}
\subfloat[$\nu_{10}$]{
\includegraphics[trim={0pt 0pt 0pt 0pt},clip, width=0.23\textwidth]{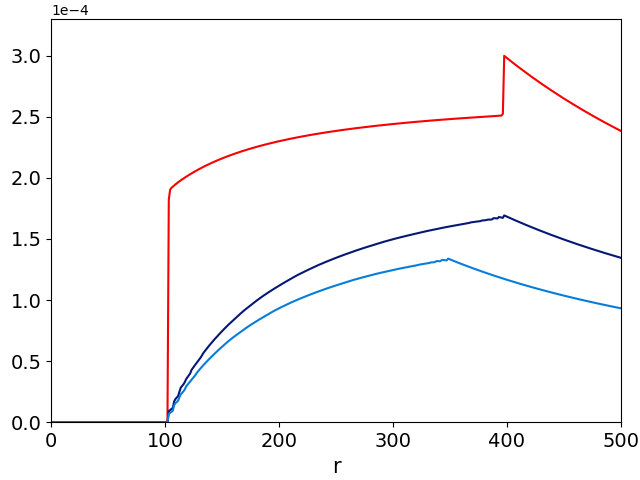}
\label{fig:subfig_sim_normed_ex_3}}
\subfloat[$\nu_{11}$.]{
\includegraphics[trim={0pt 0pt 0pt 0pt},clip, width=0.23\textwidth]{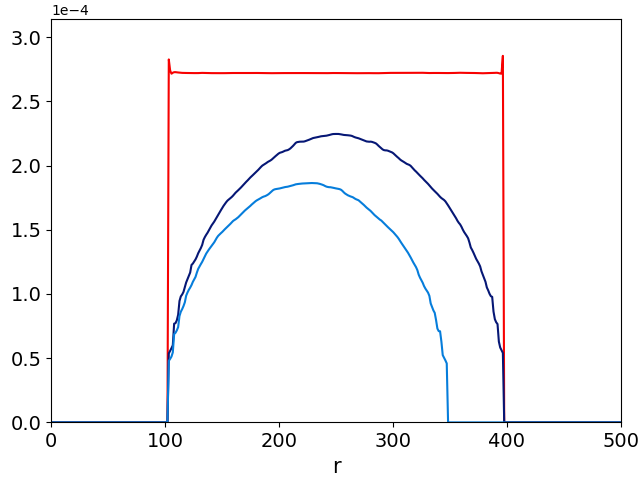}
\label{fig:subfig_sim_normed_ex_1}}
\caption{Example of measures and normalized measures. The observed objects $X$ is either of 2 spheres of different sizes and a cube as shown in Fig.~\ref{fig:subfig_sim_ex_3} and with colors corresponding to the curves. The reference object $Y$ is an infinite plane at a minimum distance of 100 units from the relevant object. The observation window is a cube of side-length 500 with a side coinciding with $Y$ and otherwise centered around the observed object.}
\label{fig:fig_sim_ex}
\end{figure}

As a first experiment and in the spirit of the analytical examples in Section~\ref{sec:analytical}, consider $\R^3$ with an infinite plane as the reference object, 2 spheres and a cube as observed objects near the plane and with a cubic observation window aligned with the reference object. Top and bottom row in Fig.~\ref{fig:fig_sim_ex} show the experimental evaluating of $\mu_{\eps,\eps'}$ and $\nu_{\eps,\eps'}$ respectively.
The experiments show that $\mu_{0,0}$ is a monotonically increasing function of the integral of the volume of the observed object from $0$ to $r$ with $\mu_{0,1}$ as its derivative. The surface measure $\mu_{1,0}$ for the spheres is a linearly increasing function, while the cube has two discontinuous steps caused by the alignment of the cube with the observation plane. The curve measure $\mu_{11}$ is quadratic for the spheres and constant for the cube. For our experimental setup, the normalisation function $N_{\eps,\eps'}$ is constant for $\nu_{\eps,1}$ giving a shape identical to $\mu_{\eps,1}$. For $\nu_{\eps,0}$ the normalization function is proportional to $r$, and thus, $\nu_{\eps,0}\sim\frac{\mu_{\eps,0}}{r}$.



\begin{figure}
\centering
\subfloat[$\mu_{0,0}(X,Y^r)$ volume measure.]{
\includegraphics[width=0.35\textwidth]{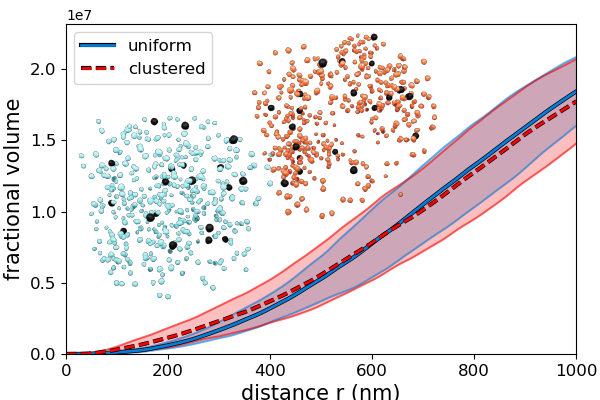}\label{fig:subfig_spheres_ex_1}}
\subfloat[$\nu_{0,0}(X,Y^r)$ normed volume.]{
\includegraphics[width=0.35\textwidth]{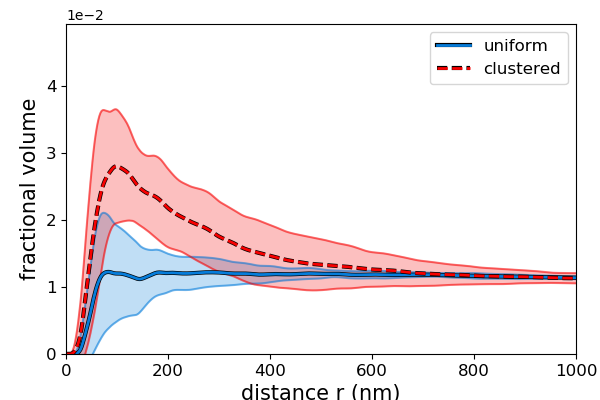}
\label{fig:subfig_spheres_ex_2}}
\caption{Example of the volume measures on uniformly distributed and clustered spheres. A slight difference can be seen directly in the $\mu_{0,0}(X,Y^r)$ graph, but the clustering is clearly visible in $\nu_{0,0}(X,Y^r)$.}
\label{fig:fig_spheres_ex}
\end{figure}

As a second synthetic experiment, we consider sets of spheres $\mathcal{X}$ and $\mathcal{Y}$ randomly distributed in a window, as shown in Fig.~\ref{fig:fig_spheres_ex}.
Looking at Fig.~\ref{fig:subfig_spheres_ex_1}, we see two $\mu_{0,0}(X,Y^r)$ volume graphs showing a slight difference between the two groups. For the normalized measure $\nu_{0,0}(X,Y^r)$, as shown in Fig.~ \ref{fig:subfig_spheres_ex_2}, we see in blue the uniformly distributed spheres show a straight line, while the clustering in the red group is shown as a peak for smaller distances of $r$. The interpretation is that the red spheres $X$ cluster mostly within distance 500 around $Y$.

\section{Experiments on cellular ultrastructures}
Communication by neurons in humans is mainly achieved by a combination of electric potential changes and chemical signaling. The vesicles serve as transient containers of the chemicals released towards another neuron at a connection point, the synapse, upon voltage potential changes in the neuron. Synaptic vesicles are therefore almost exclusively observed directly next to synapses. The mechanisms of replenishment of the vesicles are thought in part to be done by two main routes, the first is by only partly release of the neurotransmitters, the vesicle membrane is thus preserved \cite{koenig1996synaptic,fesce1994neurotransmitter}. The second is by a slow endosomatic route, where an endosome is formed from the membrane, pinched off into vesicles and filled with neurotransmitters. An open question is if the cells keep a reservoir of vesicles at some distance to the synapse with some evidence \cite{richards2000two}.
To assess the above, we examine a publicly available FIB-SEM dataset of the CA1 hippocampus region of a healthy adult rodent. Original dimensions before registration were $2048 \times 1536 \times 1065$ with a voxel size of $5\times5\times5$ nm. We have segmented the complete volumetric image into cell wall, synapses, mitochondria, vesicles, endoplasmatic reticulum, and the segmentation is available at \cite{fibsem2020annotations}.

\begin{figure}
\centering
\subfloat[An image slice.]{
\includegraphics[width=0.3\textwidth]{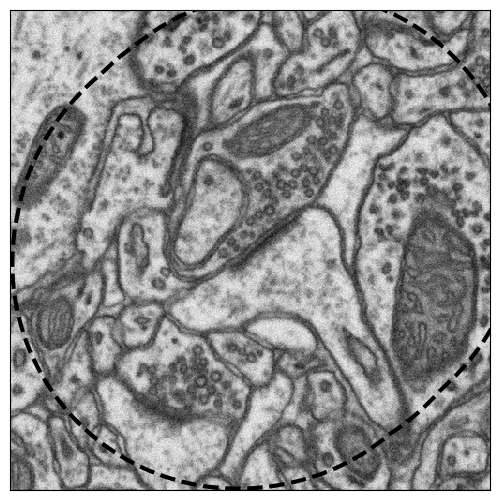}
\label{fig:subfig_lausanne_image}}
\subfloat[The slice's segments.]{
\includegraphics[width=0.3\textwidth]{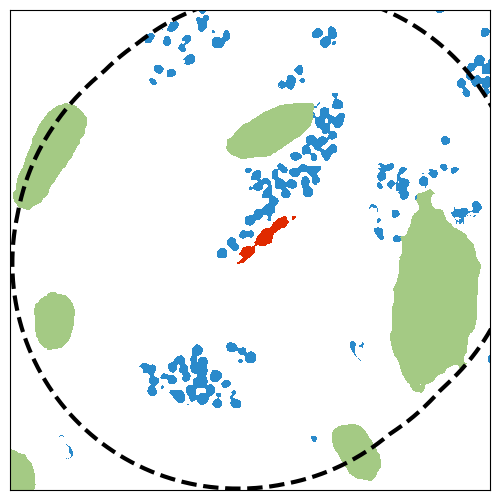}
\label{fig:subfig_lausanne_labels_2d}}
\subfloat[A 3D neighbourhood of segments.]{
\includegraphics[width=0.3\textwidth]{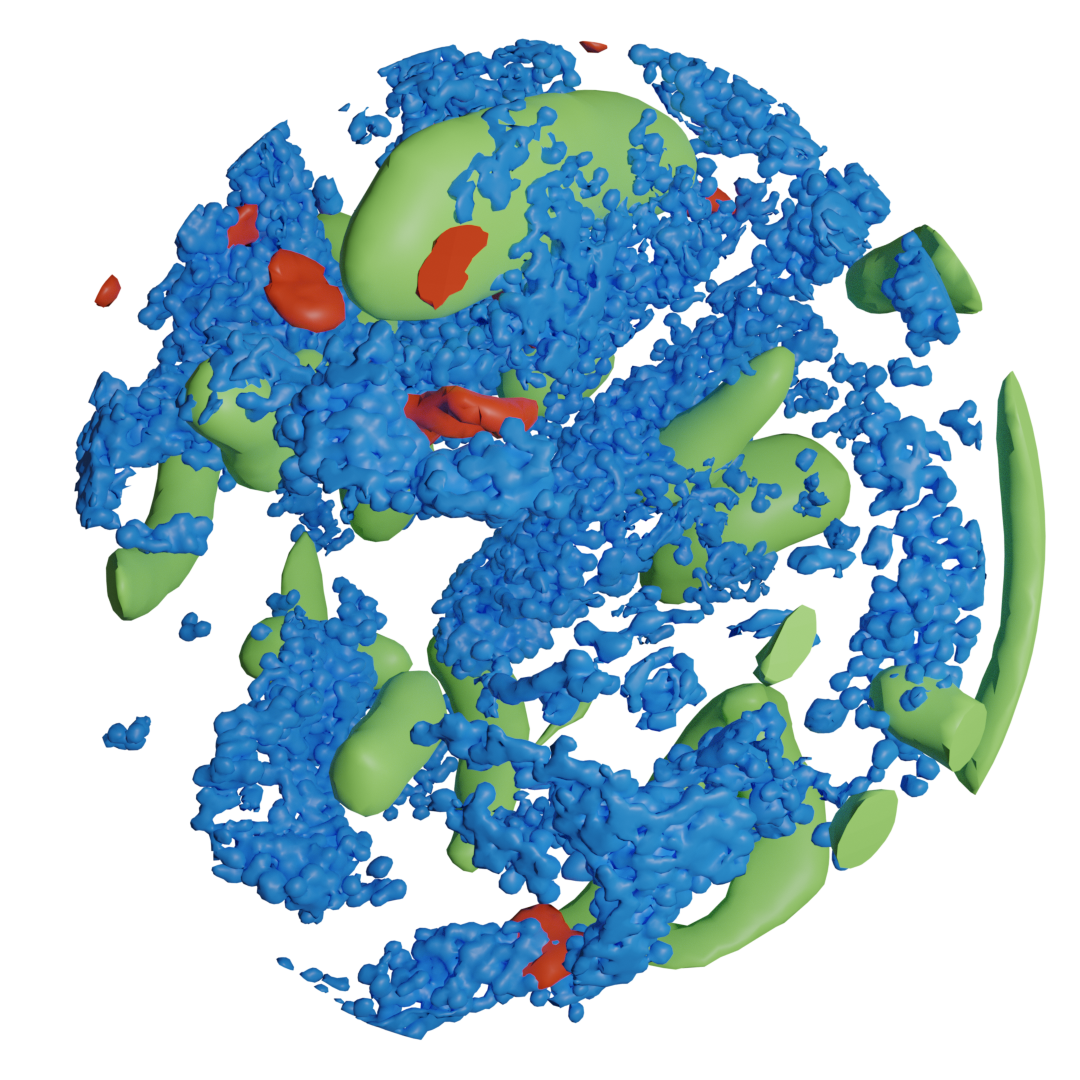}
\label{fig:subfig_lausanne_labels_3d}}
\caption{The FIB-SEM data set and our segmentation.}
\label{fig:fig_lausanne_2}
\end{figure}

The FIB-SEM were segmented by a neural network U-Net model~\cite{ronneberger2015u}, and cleaned up using an Avizo Amira pipeline~\cite{scientific2020amira}. The volume was registered to correct for drift using a model based approach described in \cite{stephensen2020restoring}. From the masks, we generate mesh reconstructions using a Marching Cubes Lewiner implementation in the SciPy Python package~\cite{2020SciPy-NMeth}. The meshing is further refined with the PyMesh Python package, and the TetWild C++ library to do mesh simplification and tetrahedralization \cite{Hu:2018:TMW:3197517.3201353}. Examples of the resulting segmentation is shown in Fig.~\ref{fig:fig_lausanne_2}

\begin{figure}
\centering
\subfloat[$L_{0,1}$ for mitochondria.]{
\includegraphics[width=0.35\textwidth]{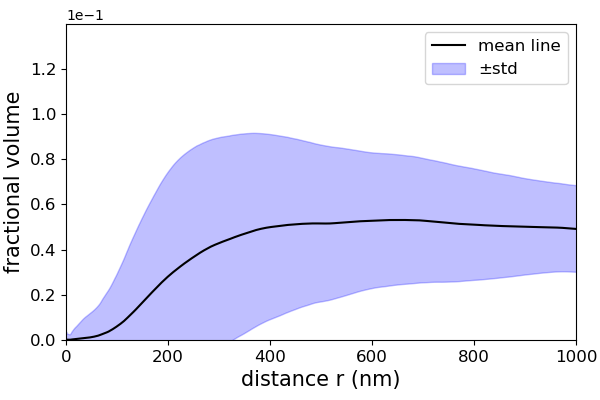}
\label{fig:subfig_lausanne_mito_normed_area}}
\subfloat[$L_{0,1}$ for vesicles.]{
\includegraphics[width=0.35\textwidth]{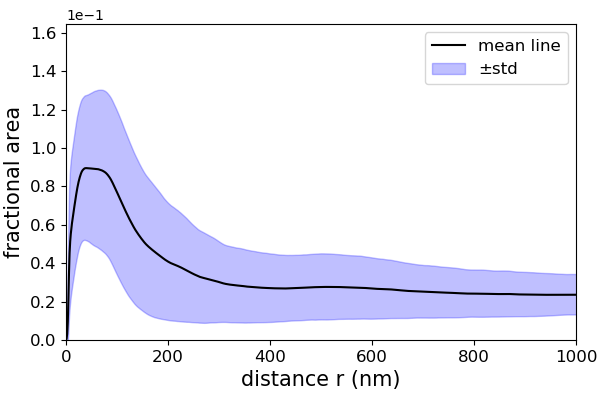}
\label{fig:subfig_lausanne_vesicle_normed_area}}
\caption{Volume comparison of vesicles and mitochondria in relation to the distance to the synapse for a total of 365 synapses and their neighbourhood vesicles and mitochondria.}
\label{fig:fig_lausanne_1_area}
\end{figure}

In Fig.~\ref{fig:fig_lausanne_1_area}, we show $L_{0,1}$ for mitochondria and vesicles when using the synapse as the reference object. No correction for cell walls have been performed.
From Fig.~\ref{fig:subfig_lausanne_mito_normed_area} we see that the mitochondria is absent close to the synapse, with a gradually increasing presence until around 800 nm, where the expected presence of mitochondria goes towards the global area fraction of mitochondria in the whole sample. From Fig.~\ref{fig:subfig_lausanne_vesicle_normed_area} we see a presence of vesicle at the very close range $\approx 25$ nm followed by a proportionally greater measure of peaking around $25-100$ nm. Before the measure tends to the global area fraction, we see a slight increase in vesicles near the 500 nm range which could indicate a presence of a vesicle reservoir at that range for these neuronal processes.

\section{Conclusion}
We have presented a novel method that extends the theoretical foundations of $K$-function summary statistics in the field of spatial point statistics to geometric objects, adds elegant reasoning about the shape of one object with respect to a reference object, and includes some existing shape-relation measures. We show the method can be used to display spatial relations such as spreading or clustering compared to uniformly random distribution, and that the method is sensitive to properties such as cross-sectional area and thickness. A core strength is that the method is built on the $n$-dimensional Hausdorff measure enabling us to intuitively understand the shape relations. Statistical tests have been developed for the comparison of $K$-functions in the context of spatial point processes~\cite{Hahn2012} and are readily available for the comparison between groups.

\bibliographystyle{splncs04}
\bibliography{refs}

\begin{thebibliography}{10}
\providecommand{\url}[1]{\texttt{#1}}
\providecommand{\urlprefix}{URL }
\providecommand{\doi}[1]{https://doi.org/#1}

\bibitem{baddeley15}
Baddeley, A., Rubak, E., Turner, R.: Spatial Point Patterns: Methodology and
  Applications with R. CRC Press (2015)

\bibitem{dixon2014r}
Dixon, P.M.: Ripley's k function. Wiley StatsRef: Statistics Reference Online
  (2014)

\bibitem{fesce1994neurotransmitter}
Fesce, R., Grohovaz, F., Valtorta, F., Meldolesi, J.: Neurotransmitter release:
  fusion or ‘kiss-and-run’? Trends in cell biology  \textbf{4}(1), ~1--4
  (1994)

\bibitem{gavrilov2018astrocytic}
Gavrilov, N., Golyagina, I., Brazhe, A., Scimemi, A., Turlapov, V., Semyanov,
  A.: Astrocytic coverage of dendritic spines, dendritic shafts, and axonal
  boutons in hippocampal neuropil. Frontiers in cellular neuroscience
  \textbf{12}, ~248 (2018)

\bibitem{Hahn2012}
Hahn, U.: {A studentized permutation test for the comparison of spatial point
  patterns}. Journal of the American Statistical Association
  \textbf{107}(498),  754--764 (2012). \doi{10.1080/01621459.2012.688463}

\bibitem{Hu:2018:TMW:3197517.3201353}
Hu, Y., Zhou, Q., Gao, X., Jacobson, A., Zorin, D., Panozzo, D.: Tetrahedral
  meshing in the wild. ACM Trans. Graph.  \textbf{37}(4),  60:1--60:14 (Jul
  2018). \doi{10.1145/3197517.3201353},
  \url{http://doi.acm.org/10.1145/3197517.3201353}

\bibitem{Khanmohammadi2015}
Khanmohammadi, M., Waagepetersen, R.P., Sporring, J.: {Analysis of shape and
  spatial interaction of synaptic vesicles using data from focused ion beam
  scanning electron microscopy (FIB-SEM)}. Frontiers in Neuroanatomy
  \textbf{9}(september) (2015). \doi{10.3389/fnana.2015.00116}

\bibitem{koenig1996synaptic}
Koenig, J., Ikeda, K.: Synaptic vesicles have two distinct recycling pathways.
  The Journal of cell biology  \textbf{135}(3),  797--808 (1996)

\bibitem{lausanne}
Lucchi, A., Li, Y., Becker, C., Fua, P.: Electron microscopy dataset.
  \url{https://cvlab.epfl.ch/data/data-em/}, accessed: 2020-03-14

\bibitem{Marques2013}
Marques, J., Genant, H.K., Lillholm, M., Dam, E.B.: {Diagnosis of
  osteoarthritis and prognosis of tibial cartilage loss by quantification of
  tibia trabecular bone from MRI}. Magnetic Resonance in Medicine
  \textbf{70}(2),  568--575 (2013). \doi{10.1002/mrm.24477}

\bibitem{medvedev2014glia}
Medvedev, N., Popov, V., Henneberger, C., Kraev, I., Rusakov, D.A., Stewart,
  M.G.: Glia selectively approach synapses on thin dendritic spines.
  Philosophical Transactions of the Royal Society B: Biological Sciences
  \textbf{369}(1654),  20140047 (2014)

\bibitem{richards2000two}
Richards, D., Guatimosim, C., Betz, W.: Two endocytic recycling routes
  selectively fill two vesicle pools in frog motor nerve terminals. Neuron
  \textbf{27}(3),  551--559 (2000)

\bibitem{Ripley1979}
Ripley, B.D.: {Tests of 'Randomness' for Spatial Point Patterns}. Journal of
  the Royal Statistical Society: Series B (Methodological)  \textbf{41}(3),
  368--374 (1979). \doi{10.1111/j.2517-6161.1979.tb01091.x}

\bibitem{ronneberger2015u}
Ronneberger, O., Fischer, P., Brox, T.: U-net: Convolutional networks for
  biomedical image segmentation. In: International Conference on Medical image
  computing and computer-assisted intervention. pp. 234--241. Springer (2015)

\bibitem{weil}
Schneider, R., Weil, W.: Stochastic and Integral Geometry. Springer, Heidelberg
  (2008)

\bibitem{scientific2020amira}
Scientific, T.F.: Amira-avizo software (2020)

\bibitem{stephensen2020restoring}
Stephensen, H.J.T., Darkner, S., Sporring, J.: Restoring drifted electron
  microscope volumes using synaptic vesicles at sub-pixel accuracy.
  Communications Biology  \textbf{3}(1), ~1--7 (2020)

\bibitem{fibsem2020annotations}
Stephensen, H.J.T., Sporring, J.: Rodent neuronal volume annotations and
  segmentations (2020),
  https://www.doi.org/10.17894/ucph.33bd30d2-5796-48f4-a0a8-96fcc0ce6af5

\bibitem{2020SciPy-NMeth}
{Virtanen}, P., {Gommers}, R., {Oliphant}, T.E., {Haberland}, M., {Reddy}, T.,
  {Cournapeau}, D., {Burovski}, E., {Peterson}, P., {Weckesser}, W., {Bright},
  J., {van der Walt}, S.J., {Brett}, M., {Wilson}, J., {Jarrod Millman}, K.,
  {Mayorov}, N., {Nelson}, A.R.J., {Jones}, E., {Kern}, R., {Larson}, E.,
  {Carey}, C., {Polat}, {\.I}., {Feng}, Y., {Moore}, E.W., {Vand erPlas}, J.,
  {Laxalde}, D., {Perktold}, J., {Cimrman}, R., {Henriksen}, I., {Quintero},
  E.A., {Harris}, C.R., {Archibald}, A.M., {Ribeiro}, A.H., {Pedregosa}, F.,
  {van Mulbregt}, P., {Contributors}, S...: {SciPy 1.0: Fundamental Algorithms
  for Scientific Computing in Python}. Nature Methods  \textbf{17},  261--272
  (2020). \doi{https://doi.org/10.1038/s41592-019-0686-2}

\bibitem{Zhang2004}
Zhang, D., Lu, G.: {Review of shape representation and description techniques}.
  Pattern Recognition  \textbf{37}(1),  1--19 (2004).
  \doi{10.1016/j.patcog.2003.07.008}

\end{thebibliography}

\end{document}